\documentclass[10pt,twocolumn,letterpaper]{article}

\usepackage{cvpr}              

\usepackage{multicol,multirow}
\usepackage{color-edits}
\addauthor{df}{cyan}
\addauthor{gn}{magenta}
\addauthor{zw}{orange}
\addauthor{jg}{blue}
\addauthor{sanjay}{green}
\addauthor{yh}{purple}
\addauthor{as}{red}

\newcommand{\data}{\textsc{SlidesBench}\xspace}
\newcommand{\lib}{\textsc{SlidesLib}\xspace}
\newcommand{\model}{\textsc{AutoPresent}\xspace}
\newcommand{\gpt}{\textsc{GPT-4o}\xspace}
\newcommand{\llama}{\textsc{LlaMa}\xspace}
\newcommand{\llava}{\textsc{LlaVa}\xspace}
\newcommand{\taskone}{\textbf{Detailed Instructions with Images}\xspace}
\newcommand{\tasktwo}{\textbf{Detailed Instructions Only}\xspace}
\newcommand{\taskthree}{\textbf{High-Level Instructions}\xspace}

\usepackage{color, colortbl}
\definecolor{amber}{rgb}{1.0, 0.75, 0.0}
\newcommand{\hlrow}{\rowcolor{amber!13}}
\definecolor{bl}{rgb}{0.738,0.875,0.992}
\newcommand{\hlrowblue}{\rowcolor{bl!40}}
\usepackage{array}

\usepackage{listings}

\definecolor{cvprblue}{rgb}{0.21,0.49,0.74}
\usepackage[pagebackref,breaklinks,colorlinks,allcolors=cvprblue]{hyperref}

\def\paperID{17296} 
\def\confName{CVPR}
\def\confYear{2025}

\title{\model: Designing Structured Visuals from Scratch}

\author{
  Jiaxin Ge\textsuperscript{1}\footnotemark[1] \quad Zora Zhiruo Wang\textsuperscript{2}\footnotemark[1] \\ Xuhui Zhou\textsuperscript{2} \quad Yi-Hao Peng\textsuperscript{2}\quad Sanjay Subramanian\textsuperscript{1}
  \quad Qinyue Tan\textsuperscript{2}\\ Maarten Sap\textsuperscript{2} \quad Alane Suhr\textsuperscript{1}\footnotemark[2]\quad Daniel Fried\textsuperscript{2}\footnotemark[2]\quad Graham Neubig\textsuperscript{2}\footnotemark[2] \quad
  Trevor Darrell\textsuperscript{1}\footnotemark[2]\\ \\
  \textsuperscript{1}University of California, Berkeley \quad \textsuperscript{2}Carnegie Mellon University\\
}

\usepackage{cuted}
\usepackage{capt-of}
\usepackage{float}

\begin{document}

\maketitle
\footnotetext[1]{Equal Contribution.}
\footnotetext[2]{Equal Contribution.}
\begin{strip}\centering
\vspace{-12px}
\includegraphics[width=0.99\textwidth]{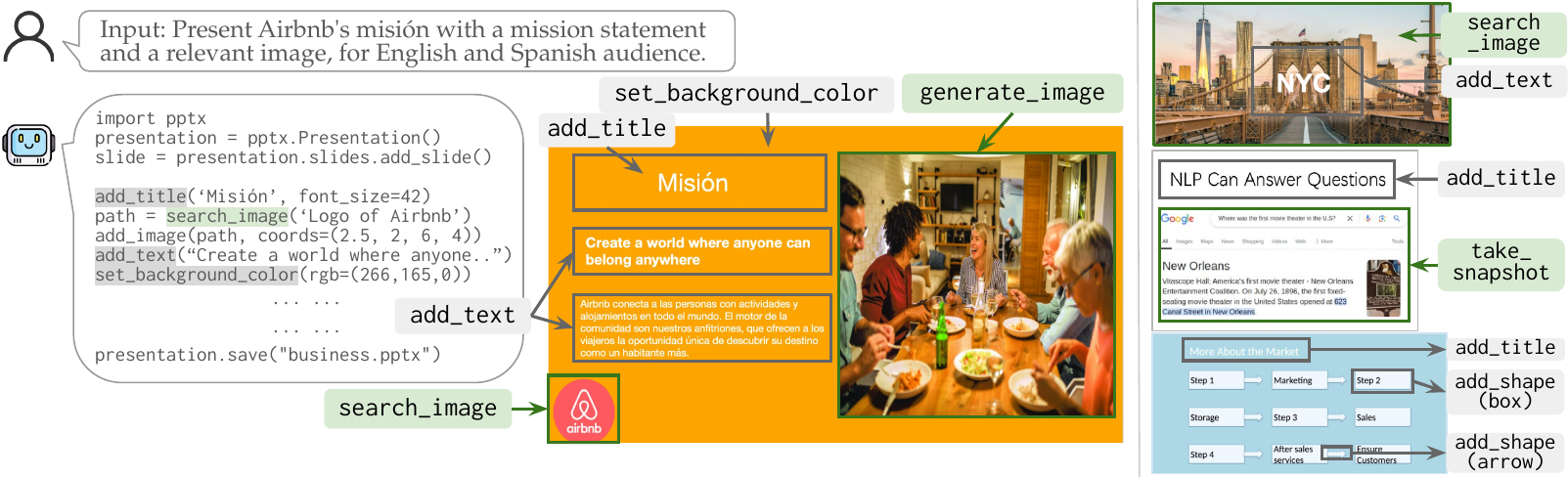}
\vspace{2mm}
\captionof{figure}{\textbf{Automatically generating slides from natural language instructions.} We propose \model, a tool-augmented code generation method that follows natural language instructions to design slides from scratch, as shown in the examples. This allows for precise control over all elements, including textual content, images, visual layouts, coloring, and more.
}
\label{fig:teaser}
\end{strip}

\begin{abstract}
Designing structured visuals such as presentation slides is essential for communicative needs, necessitating both content creation and visual planning skills.
In this work, we tackle the challenge of automated slide generation, where models produce slide presentations from natural language (NL) instructions.
We first introduce the \data benchmark, the first benchmark for slide generation with 7k training and 585 testing examples derived from 310 slide decks across 10 domains. 
\data supports evaluations that are (i) {\it reference-based} to measure similarity to a target slide, and (ii) {\it reference-free} to measure the design quality of generated slides alone.
We benchmark end-to-end image generation and program generation methods with a variety of models, and find that programmatic methods produce higher-quality slides in user-interactable formats. 
Built on the success of program generation, we create \model, an 8B \llama-based model trained on 7$k$ pairs of instructions paired with code for slide generation, and achieve results comparable to the closed-source model \gpt.
We further explore iterative design refinement where the model is tasked to self-refine its own output, and we found that this process improves the slide's quality.
We hope that our work will provide a basis for future work on generating structured visuals. Our code, data, demo, and video demonstrations are
publicly available at \url{https://github.com/para-lost/AutoPresent}
\end{abstract}

\vspace{-15pt}    
\section{Introduction}
\label{sec:intro}

Designing structured visuals such as presentation slides from scratch is an essential skill for effective communication and conveying complex ideas \citep{reynolds2011presentation}. 
Among various forms of visual communication, creating a compelling set of slides is a challenging problem, requiring content creation (text, pictures, diagrams, and more) and visual planning skills, to ensure the slides are well designed \citep{kong2022aesthetics} and convey insights with clarity \citep{al2005auto,sefid2019automatic}.
Even human experts may need to spend hours iterating and polishing their slide decks \citep{decktpus} to produce high-quality designs with clear insights. 
While digital agents have demonstrated impressive capabilities in tasks such as
software engineering \cite{yang2024swe}, web navigation \citep{zhou2024webarena,wang2024agent}, and free-form image design generation \citep{betker2023improving, rombach2022high}, their creative capabilities in generating semi-structured communicative media like slide decks has not been extensively tested. Therefore, we ask: \textit{Can we employ powerful AI agents to create high-quality presentation slides that are well-structured and insight-revealing?}

In this work, we formulate the natural language (NL) to slide generation task. At a high level, the user provides the system with a natural language instruction about the desired slide, and the system then generates an editable presentation, as shown in \autoref{fig:teaser}. 
We consider three types of user instructions:
(1) {\it detailed instruction with images}. (2) {\it detailed instructions only}. (3) {\it high-level instructions}, reflecting varying levels of design freedom.

Since there are no existing tools for quantifying agent performance in slide generation tasks, we propose the \data benchmark (\S\ref{sec:2:slides-bench}) as a training source and test bed for method comparisons.
\data contains 7$k$ training examples and 585 testing examples of varied instruction difficulties, constructed from 310 publicly available slide decks from 10 different domains, including art, business, and technology.
To evaluate generated slides, we introduce
two sets of evaluation metrics: {\it reference-based} metrics to examine position, content, and color match against the reference slide; and {\it reference-free} metrics inspired by slide design principles \citep{linkedin,piktochart,24slides,four,latest} to measure the design quality of agent-created slides alone, given that many good designs for the same instructions may vary from the reference slide. 

To enable controlled and structured slide generation, we propose to create slides using program generation, where a model first generates a program from the natural language instruction, and then the program is executed to get the slide.
We apply this approach to large language models (LLMs; \llama~\citep{dubey2024llama}, DeepseekCoder~\citep{guo2024deepseek}, CodeLlama~\citep{roziere2023code}, \gpt~\citep{achiam2023gpt}) and vision-language models (VLMs; \llava~\citep{yang2024posterllava}). As illustrated in \autoref{fig:teaser}, given a natural language instruction, the model first generates a Python program and then executes it to obtain a PPTX slide.
We find that small models such as \llama (8B) and \llava (7B) are often unable to produce executable code. While GPT-4o can produce reasonable slides, it still exhibits a substantial gap in design quality compared to human-generated slides (\S\ref{sec:4:results}). 
By further conducting iterative refinement, we find that models can self-refine and further improve slide quality. We also find that code generation approaches substantially outperform end-to-end image generation methods (Stable Diffusion~\cite{rombach2022high}, Dall-E~\cite{betker2023improving}).

To further enhance the current model's ability to generate high-quality slides,
we present our open-sourced \model (8B) model (\S\ref{sec:3.2:slides-llama}) which is fine-tuned from \llama 8B on the \data training set. \model achieves state-of-the-art performance among small open-sourced models and approaches the performance of the closed-sourced model GPT-4o.
Since directly generating a long program is difficult for current models~\citep{ge2025recursive}, we further create the \lib library to simplify the program generation process. \lib contains high-level functions that are {\it basic} such as \texttt{add\_title}, and {\it image-related} such as \texttt{search\_image} and \texttt{generate\_image}. We show that LLMs and VLMs generally perform better when given access to \lib.

Our main contributions can be summarized as follows:
\begin{itemize}
    \item We formulate the NL-to-slide generation task and build \data, the first benchmark for slide generation, which contains 7$k$ training and 585 test examples and supports automatic evaluations.
    \item We leverage NL-to-program generation methods with refinement to produce high-quality slides, and benchmark diffusion models, VLMs, and LLMs.
    \item We train an 8B parameter open-source LLM, \model, that approaches the performance of GPT-4o, and design a programmatic tool library \lib that facilitates slide program generation across models.     
\end{itemize}

\begin{figure*}[t!]
\vspace{-3mm}
    \centering
    \vspace{-1mm}
    \includegraphics[width=\textwidth]{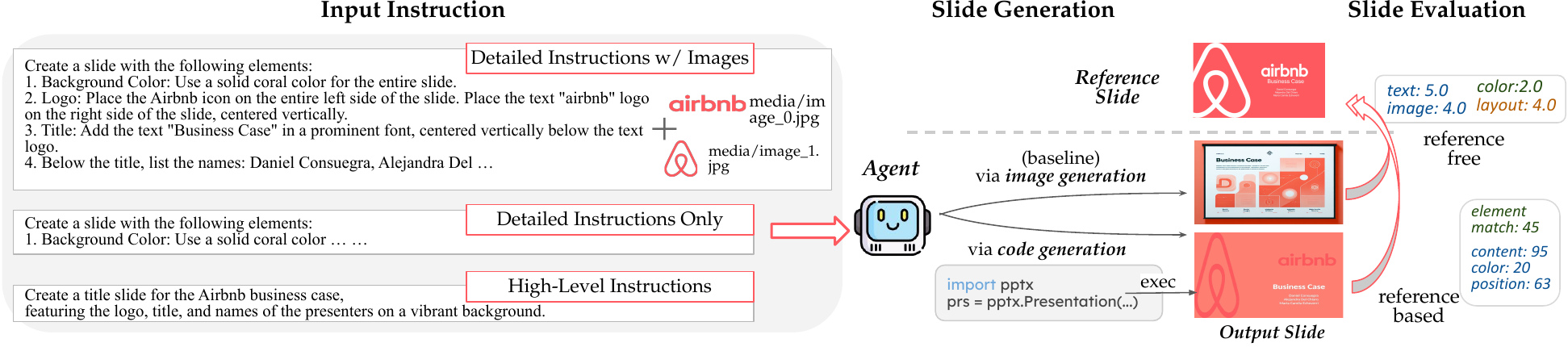}
    \vspace{-4mm}
    \caption{\textbf{Illustration of \data.} Each example of \data consists of three instructions: Detailed Instructions with Images, Detailed Instructions Only, and High-Level Instructions. The model is tasked to generate a slide based on the instruction, and the generated slide is evaluated on the metrics suite, which contains both the reference-free metrics and the reference-based metrics.}
    \label{fig:benchmark-overview}
    \vspace{-2mm}
\end{figure*}

\section{\data}
\label{sec:2:slides-bench}

In this section, we describe the creation of the \data benchmark. Each instance consists of a natural language instruction to create a slide, and the slide itself (in PPTX format) as a reference. \data includes three scenarios of varying difficulty levels designed to evaluate models with different user input.
We describe the slide data collection (\S\ref{sec:2.1:data-collect}), three task setups (\S\ref{sec:2.2:three-challenges}), and the annotation process (\S\ref{sec:2.3:example-annotation}).

\subsection{Slides Data Collection}
\label{sec:2.1:data-collect}
We search the web and collect presentation slide decks from 10 domains, including art, marketing, environment, technology, etc.
To select the highest-quality slide decks from each domain, we manually go through the relevant slide decks and conduct initial processing, by checking if all its slides (i) have visually structured layouts, and (ii) extractable media such as images (if any). For the slide decks with all slides satisfying (i) and (ii) in each domain, we incorporate one slide deck into the test set, and others into the training set.
This results in a total of 10 and 300 slide decks (in \texttt{PPTX} format) for testing and training, each containing 20 slides on average.
To respect the rights of the slide creators, we do not redistribute the slides. Instead, we provide a list of URLs for the slides that we used so that others can download the slides directly from the original website. We also provide an opt-out mechanism for any creator who does not want their slides in the dataset.
We provide implementation details in \S\ref{app:implementation}.

\subsection{Three Task Setups}
\label{sec:2.2:three-challenges}
We formulate the task as an NL-to-slide generation process. Given the reference slide, we curate three versions of natural language instructions, as shown in \autoref{fig:benchmark-overview}, to represent slide generation tasks under varied difficulty levels. We introduce each setup below.

\noindent \taskone \quad
The first and easiest setting is to provide the models with all the necessary information and assets to produce the reference slide, including text and image content, formatting and layout specifications. This setting evaluates models' visual planning abilities, such as arranging spatial layouts, maintaining formatting consistency, balancing content proportions, and emphasizing key elements.

\noindent \tasktwo \quad
Since a user may not specify, or know exactly what images to put on a slide, we propose a \textit{detailed instruction only} setting, where we provide the same natural language instruction provided in the \textit{detailed instruction with images} setting, but replace the provided images with their natural language descriptions (e.g., ``two people shaking hands'') generated by \texttt{gpt-4o-mini}. We then instruct the models to obtain the images using image searching or image generation tools. 
This setting further challenges models to interpret complex or compositional descriptions of images and obtain visuals that align with the slide context.

\noindent \taskthree \quad
In contrast to users who have a concrete target slide in mind and can spell out all detailed instructions, some users may only be able to express their needs on a high level. We thus devise a \textit{ high-level instruction} setting, where the natural language instructions are rather high-level and only provide a general topical idea of the slide, such as ``create a title slide for Airbnb,'' instead of detailing what logos and text to add and where, as exemplified in \autoref{fig:benchmark-overview}.
Models in this case need to both acquire or create content, and arrange the elements properly.

\subsection{Example Annotation}
\label{sec:2.3:example-annotation}
To annotate the dataset, we collect natural language instructions paired with each slide.
For each slide, we create three versions corresponding to the three setups in \S\ref{sec:2.2:three-challenges}.

\noindent \taskone \quad
To produce detailed instruction with images, we use a scalable approach combining human-written examples and model-generated annotations. 
For each slide deck, we first write instructions for three example slides manually --- including all necessary information (content, layout, formatting) to reproduce the slide, and providing paths to the images used in the slide (e.g., \texttt{media/image\_0.png}), as shown in \autoref{fig:benchmark-overview} (top).
We then use these (human-written instruction, reference slide) pairs as few-shot examples to prompt LLM (specifically, \texttt{gpt-4o-mini}) to generate natural language instructions for each slide in the current slide deck.\footnote{Including the three slides with human-written instructions, to ensure instructions for all slides are consistent in style and specificity.} 
Further, for the test set, we manually examined and refined the instructions by correcting incorrect specifications, adding missing content, and removing unnecessary or untrue content.

\noindent \tasktwo \quad
To produce detailed instruction only, we 
replace the image paths (e.g., \texttt{media/image\_0.png}) with the natural language descriptions of the images(\texttt{``an artistic, colorful background''}). 
These descriptions are generated by \texttt{gpt-4o-mini}. 
For the test set, we manually refine the instructions to ensure that they do not refer to unavailable image paths (e.g., removing phrases like ``use the provided images"), as shown in \autoref{fig:benchmark-overview} (middle). 

\noindent \taskthree \quad
To create high-level instructions, we start with a similar approach by manually annotating three examples and then prompting the model to generate for all slides. Human-written instructions only provide a topical description of the slide and intentionally leave out specific content or layout details.
This process ensures that the generated instructions remain concise and general, as shown in \autoref{fig:benchmark-overview} (bottom).

Overall, the instructions have an average of 115.6, 118.3, and 26.6 words under {\it detailed instruction with images}, {\it detailed instruction only}, and {\it high-level instructions} settings respectively, accompanied by an average of 1.1, 0.0, and 0.0 provided images.


\section{Evaluation Metrics}
\label{sec:eval-metrics}
In this section, we describe the evaluation metrics that we designed for \data.
We propose two sets of evaluation metrics: reference-based metrics for measuring models' instruction-following abilities (\S\ref{sec:eval-metrics_ref_based}), and reference-free metrics to examine the design quality of model-generated content (\S\ref{sec:eval-metrics_ref_free}). We also use executability to examine the success rate of each model (\S\ref{sec:eval-metrics_exe}).

\subsection{Reference-Based Metrics}
\label{sec:eval-metrics_ref_based}
Inspired by Design2Code \citep{si2024design2code} metrics, we implement four dimensions to examine the similarity between the model-produced slides and the reference slide.

\noindent {\bf Element matching} \quad
For the slide layout, we measure the total sizes of matched elements (in generated and reference slides) divided by the total sizes of all element, where each textbox, image, or shape constitutes an element.
More concretely, we accurately parse out each element in the generated and reference slides, and compute their maximum matching using the \texttt{match} library.

\noindent {\bf Content similarity} \quad
For each pair of matched elements, we compute their content similarity. If the reference element is text, we calculate the textual similarity using the cosine similarity of the embeddings produced with sentence-transformer with the default \texttt{all-MiniLM-L6-v2} model \citep{reimers-2019-sentence-bert}.
If the reference element is an image, we calculate the CLIP score \citep{hessel2021clipscore} of the image in two elements.
We report the average content similarity across all matched element pairs, if either element contains a non-empty text string or an image component.

\noindent {\bf Color similarity} \quad
We also measure the coloring similarity using the CIEDE2000 color difference formula \citep{luo2001development}, to quantify the perceptual difference between the colors. For every matched element pair, we measure the text font color similarity and element background color (if any). We additionally measure the color similarity between the background color of generated and reference slides. 

\noindent {\bf Position similarity} \quad
In addition to content and formatting, we also calculate the positional similarity between each pair of matched elements. More concretely, we follow \citet{si2024design2code} to normalize the element coordinates to $[0,1]$ by the slide page length and width. We compute the Manhattan distance between the elements and formulate positional similarity as $sim(r, g) = 1 - max(abs(x_r-x_g, y_r-y_g))$.

Note that a low text, color, or position similarity score could come from differences in text, color, and positions, or derivative errors caused by the inaccurate element-matching process (e.g., it may match the title box in the generated slide to a content textbox in the reference slide, which has different content or coloring requirements).

\subsection{Reference-Free Metrics}
\label{sec:eval-metrics_ref_free}
A well-designed slide generated by models may look very different from the reference slide. Therefore, we also propose four reference-free evaluation metrics, to independently assess the design quality of model-generated slides.
To establish the metrics, we surveyed a wide range of literature on slide design principles \citep{linkedin,piktochart,24slides,four,latest}, and summarized four major points as below and detailed in \autoref{tab:ref-free-metrics}:

\begin{table}[ht]
\centering
\small
\resizebox{\columnwidth}{!}{
\begin{tabular}{cp{0.40\textwidth}} 
\toprule 
{\bf Metric} & \multicolumn{1}{c}{\bf Criteria} \\
\midrule
{Text} & {The title should be simple and clear to indicate the main point. For main content, avoid too many texts and keep words concise. Use a consistent and readable font size, style, and color.} \\
\midrule
{Image} & {Use high-quality images with a reasonable proportion.} \\
\midrule
{Layout} & {Elements should be aligned, do not overlap, and have sufficient margins to each other. All elements should not exceed the page.} \\
\midrule
{Color} & {Use high-contrast color especially between the text and the background. Avoid using high-glaring colors.} \\
\bottomrule 
\end{tabular}
}
\vspace{-2mm}
\caption{Reference-free metrics, all evaluated in 0-5 scale.}
\vspace{-2mm}
\label{tab:ref-free-metrics}
\end{table}

\noindent {\bf Text} \quad
Using concise texts is important for slides to engage with the audience. An ideal slide should have a clear title, concise main content, and readable formatting.

\noindent {\bf Image} \quad
Using appropriate visuals can engage audiences. We hence measure if models can find high-quality images and properly use them to enhance the slide quality.

\noindent {\bf Layout} \quad
Slide layout is crucial to create visual balance. We examine whether all elements are within the slide, have no overlap, and align properly with the relevant elements.

\noindent {\bf Color} \quad
Vivid and consistent color use in slides can help deliver insights. We check if the slide uses high-contrast colors to facilitate visibility, and avoid high-glaring colors to discourage user engagement.

\paragraph{Validation of Reference-Free Evaluation}
For all the metrics, we provide the image version of the slide and ask the \texttt{gpt-4o} model to produce a score between 0--5.
To examine the reliability of this model-based evaluation, we conduct a human study and compare the intraclass correlation coefficient (ICC) between two human annotators and model evaluation, on all ground-truth slides.
Our examination gives high ICC scores across all four metrics: $73.8\%$--$85.3\%$, which are well within the range of what is typically considered ``high agreement''. 
In experiments in later sections, we scale these 0-5 scale scores to the 0--100 range to enable comparisons on this more standard scale.
\footnote{We still evaluate with 0-5 scale to maintain a robust, human-aligned evaluation process.}
\subsection{Executability}
\label{sec:eval-metrics_exe}
Particularly for methods based on code generation (\S\ref{sec:3.1:code-gen}), we additionally measure the execution success rate to account for invalid programs. Concretely, we count the percentage of successfully executing programs generated by models among all examples.
We report reference-based and free scores for executable slides only, to fairly compare their design quality. But we report `Overall' scores for all slides by assigning zeros to non-executing slides, to account for execution failures.
We report all metrics for successfully executing and all slides in \S\ref{app:results}.
\section{Method}
\label{sec:3:method}

We introduce our main method --- slide generation via code generation, optionally using our \lib toolkit (\S\ref{sec:3.1:code-gen}). 
Then, we present \model, trained on 7$k$ slides, that achieves performance on par with strong GPT model  (\S\ref{sec:3.2:slides-llama}).

\subsection{Slides via NL-to-Code Generation}
\label{sec:3.1:code-gen}
\paragraph{Generating Python Programs}
Given natural language instructions in \S\ref{sec:2:slides-bench}, models are tasked with generating Python programs using publicly available libraries such as \texttt{python-pptx}.
The model receives two (natural language instruction, Python program) pairs as in-context examples, followed by the test instruction, and generates a Python program which is then executed and will ideally yield a \texttt{PPTX} file containing the requested slide.
\paragraph{Generating Programs with \lib}
Nonetheless, the programs above could be very long and complex (170 lines on average), which could be challenging for models to generate entirely correctly, as shown in previous work \citep{ge2025recursive}. To address this, we design \lib, a library that provides easier-to-use interfaces for several common actions such as setting a title or setting background color.
Using \lib, the average program length is reduced to 13 lines, significantly easing the generation task.
As shown in \autoref{tab:our-library}, \lib includes 4 functions for basic operations and 3 functions for image search and generation, these functions allow models to produce more concise and modular programs. 
To enable the model to generate programs using \lib, we follow the visual programming method \citep{suris2023vipergpt} by providing a prompt that includes the documentation of the functions and two in-context examples. See more \lib details in \S\ref{app:library}.

\begin{table}[ht]
\vspace{-1mm}
\small
\centering
\begin{tabular}{p{0.18\textwidth}|p{0.27\textwidth}} 
\toprule 
\multicolumn{1}{c|}{\bf Function} & \multicolumn{1}{c}{\bf Description} \\
\midrule
\texttt{add\_title} & {Insert a title in the slide.} \\
\texttt{add\_text} & {Insert text at a specific location.} \\
\texttt{add\_bullet\_points} & {Insert a textbox with bullet points.} \\
\texttt{add\_image} & {Insert image at a specific location.} \\
\midrule
\texttt{generate\_image} & {Call an image generator (Dall-E 3) given a query.} \\
\texttt{search\_image} & {Search for an image on a search engine (Bing).} \\
\texttt{search\_screenshot} & {Display a query on a web browser (Google Chrome) and take a snapshot of the search result.} \\
\bottomrule 
\end{tabular}
\vspace{-2mm}
\caption{Basic (top) and image-specific (bottom) functions provided by \lib.}
\label{tab:our-library}
\vspace{-4mm}
\end{table}

\begin{table*}[t!]
\centering
\small
\vspace{-3mm}
\resizebox{0.85\textwidth}{!}{
\renewcommand{\arraystretch}{1.2} 
\begin{tabular}{l|c|cccc|cccc|c} 
\toprule 
\multicolumn{1}{c|}{\multirow{2}{*}{\bf Method}} & \multirow{2}{*}{\bf Execution\%} & \multicolumn{4}{c|}{\bf Reference-Based} & \multicolumn{4}{c|}{\bf Reference-Free}  & \multirow{2}{*}{\bf Overall} \\ 
\multicolumn{1}{c|}{} & {} & {\bf element} & {\bf content} & {\bf color} & {\bf position} &{\bf text} & {\bf image}& {\bf layout} & {\bf color} & {} \\
\midrule
{Reference} & {100.0} & \multicolumn{4}{c|}{--} & {59.7} & {81.5} & {73.5} & {65.7} & -- \\ 
\midrule 
\hlrow \multicolumn{11}{c}{\it End-to-end Image Generation} \\ 
\midrule 
{Stable-Diffusion*} & {\bf 100.0} & {74.5} & {33.4} & {9.0} & {75.0} & {19.6} & {45.1} & {36.9} & {40.5} & {48.0} \\ 
{DALLE 3*} & {\bf 100.0} & {75.5} & {39.9} & {9.2} & {76.1} & {32.7} & {87.3} & {56.7} & {53.4} & {50.2} \\ 
\midrule 
\hlrow \multicolumn{11}{c}{\it Code Generation w/o \lib} \\ 
\midrule 
{LLaVA (7B)} & {11.3} & {61.9} & {\bf 97.3} & {$~~$6.2} & {70.8} & {41.6} & {\bf 100.0} & {29.2} & {25.7} & {$~~$6.1} \\ 
{CodeLLaMA (7B)} & {$~~$5.1} & {63.6} & {94.0} & {11.2} & {74.0} & {52.0} & {43.0} & {48.0} & {40.0} & {$~~$3.1} \\ 
{LLaMA (8B)} & {$~~$2.1} & {74.0} & {94.6} & {12.5} & {\bf 81.2} & {50.0} & {$~~$8.3} & {50.0} & {50.0} & {$~~$1.3} \\
{GPT-4o} & {89.2} & {83.3} & {91.6} & {10.5} & {77.0} & {51.9} & {72.8} & {53.7} & {54.7} & {55.1} \\ 
\midrule
{\textbf{\model} (ours)} & {79.0} & {67.7} & {79.7} & {10.9} & {75.9} & {45.3} & {$~~$62.7} & {54.2} & {60.9} & {45.2} \\ 
\midrule
\hlrow \multicolumn{11}{c}{\it Code Generation w/ \lib} \\ 
\midrule 
{LLaVA (7B)} & {20.0} & {80.5} & {80.5} & {$~~$3.5} & {64.0} & {37.5} & {48.0} & {29.5} & {43.5} & {$~~$9.7} \\ 
{CodeLLaMA (7B)} & {48.7} & {80.3} & {89.8} & {9.4} & {69.3} & {45.9} & {66.8} & {45.1} & {49.9} & {30.3} \\ 
{LLaMA (8B)} & {54.4} & {78.3} & {91.2} & {7.5} & {69.5} & {46.0} & {68.2} & {47.6} & {53.1} & {33.5} \\ 
{GPT-4o} & {86.7} & {\bf 86.2} & {92.5} & {12.7} & {76.3} & {\bf 54.6} & {83.7} & {\bf 70.5} & {59.4} & {\bf 58.0} \\ 
\midrule
{\textbf{\model} (ours)} & {84.1} & {84.2} & {92.2} & {\bf 18.1} & {67.2} & {47.8} & {73.2} & {58.6} & {\bf 64.7} & {55.0} \\ 
\bottomrule 
\end{tabular}
}
\vspace{-1mm}
\caption{\textbf{Results with \textit{detailed instructions with images}.} We found that small models like \llava (7B) and \llama (8B) can barely generate any slides, while \model (8B) generates slides on par with GPT-4o. All the models still underperform humans. 
}
\label{tab:results-sufficient}
\vspace{-2mm}
\end{table*}
\subsection{\textbf{\large \model}}
\label{sec:3.2:slides-llama}

Using the slides in the training set of \data, we construct (natural language
instruction, program) pairs to form training data to train an open-sourced 8B model, \model. This model is based on the \llama-3.1-8B-Instruct and trained using LoRA \citep{hulora} with a rank of 128.
\paragraph{Training Data Construction}
To create (natural language instruction, program) training pairs, we generate two versions of canonical program solutions for each slide: 

\noindent \textbf{(i) Basic Python Programs} \quad
We derive canonical programs (that is, programs that can be executed to reproduce the slide) without \lib. 
To do this, we manually design an extraction script that (i) extracts each element (e.g., text and image) on the given slide, and (ii) produces a rule-based program that adds each element to the slide.
After extracting and adding each element to the slide, the resulting program accurately reproduces the original slide. 

\noindent \textbf{(ii) \lib Python Programs} \quad
We also generate canonical programs using \lib, by transforming snippets from the programs above into \lib function calls. To reproduce images in \textit{detailed instruction only} and \textit{high-level instructions} settings, we generate a short caption for each image and provide it to GPT-4o to generate the program for producing that image using \texttt{search\_image} or \texttt{generate\_image} functions.
More details of this automatic program generation process are in \S\ref{app:lib:example-programs}.
\paragraph{Training Set Composition}
After obtaining three instructions and two program versions for each example, we construct four versions of the training data, each with 7$k$ examples: 
\begin{enumerate}
\small
    \item \texttt{(detailed instruction with images, python program)}
    \item \texttt{(detailed instruction with images, \lib program)} 
    \item \texttt{(detailed instruction only, \lib program)} 
    \item \texttt{(high-level instructions, \lib program)} 
\end{enumerate}
These training sets allowed us to train four specialized models that address different challenges, which we report in \autoref{tab:results-sufficient} (1,2) and \autoref{tab:results-others} (3,4).

\subsection{Iterative Refinement}
\label{sec:4.3:refine}
Slide generation is by nature an iterative process and often requires visual-based refinements after the first draft.
To enable models to refine slides as humans do, we explore an iterative refinement procedure, where the model is tasked to self-refine the slide it generated.
Specifically, in the setting using \lib, we provide GPT-4o (capable of consuming slide images) with the original language instruction, the program it generated in the first pass, and a snapshot of the rendered slide; the model is then asked to generate a new program based on these information to refine the slide quality by tweaking colors, spacing, and other aspects of the slide. See the prompts of this process in \S\ref{app:refinement}.
\section{Experiments and Results}
\label{sec:4:results}
\begin{figure*}[h!]
    \centering
    \includegraphics[width=0.95\textwidth]{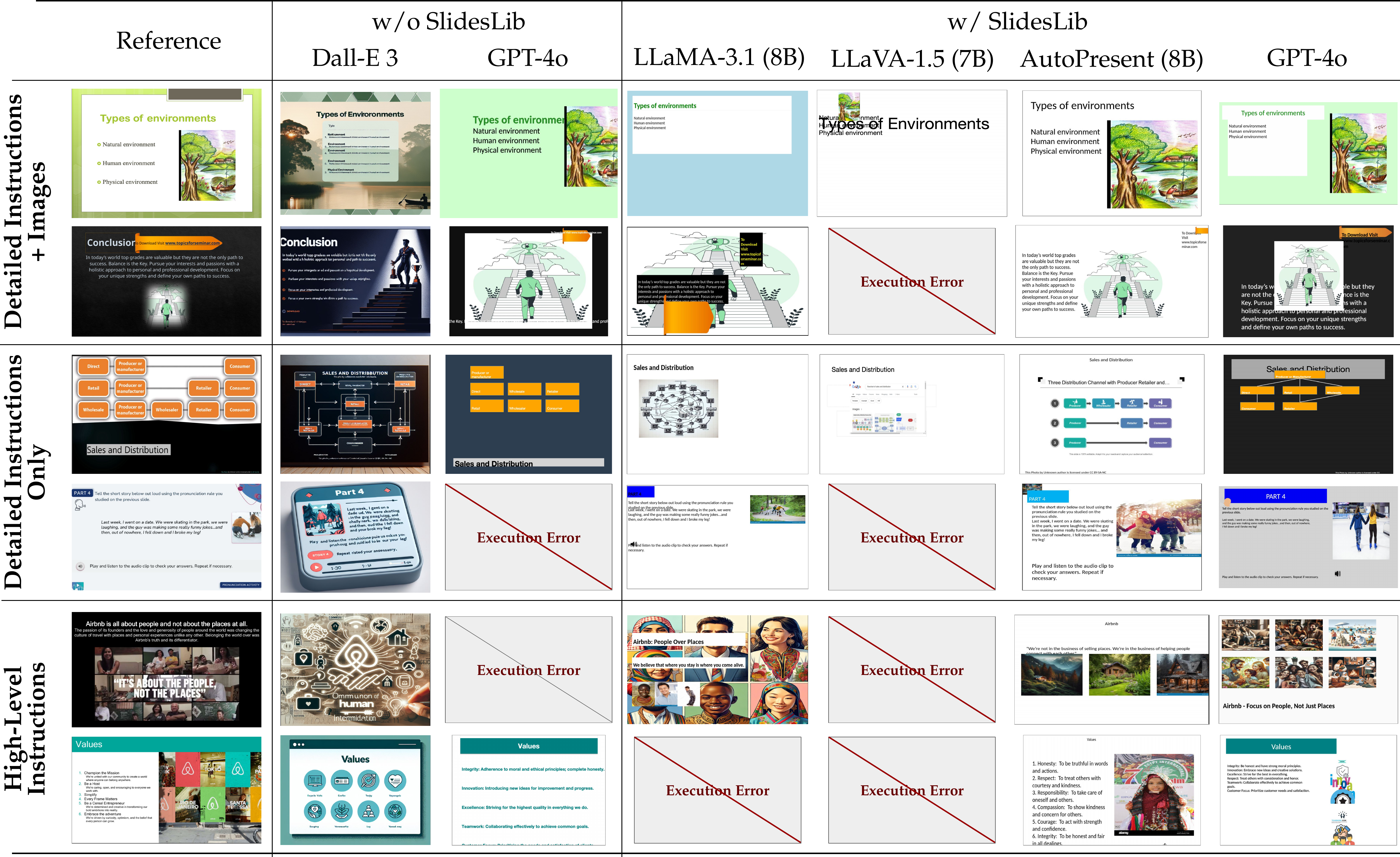}
    \caption{\textbf{Examples of slides generated by different methods in three scenarios.} End-to-end image generation methods fail to generate structured and clear slides. Small open-sourced models like \llama and \llava can barely generate any usable slides, while \model produces quality slides. Adding \lib improves GPT-4o's performance on \textit{detailed instruction only} and \textit{high-level instruction} tasks.
    }
    \label{fig:qualitative-examples}
\end{figure*}

We first introduce the experimental setup (\S\ref{sec:4.1:expr-setup}), then present the results under various scenarios (\S\ref{sec:4.2:results}).

\subsection{Experimental Setup}
\label{sec:4.1:expr-setup}
\paragraph{Code Generation Approaches}
For code generation approaches, we sample $n=3$ responses and iteratively go through them, using the first successfully executing program as the final output of the model. If none of the $n$ responses execute successfully, we count it as an execution failure.
In addition to \model (\S\ref{sec:3.2:slides-llama}), 
we benchmark several LMs out-of-the-box, including open-weights \textsc{Llama 3.1} (8B, Instruct), the code generation models DeepseekCoder-7B-v1.5 and CodeLlaMa-7B-Instruct, the vision-language \llava v1.5 model (with a Vicuna-7B-v1.5 LM backbone); and the proprietary \gpt model (the \texttt{gpt-4o-2024-08-06} checkpoint).

\paragraph{End-to-End Image Generation} 
We compare code generation with end-to-end neural image generation methods, which are a common way to produce visuals. These methods are good at creating scenic or artistic images, but may be imprecise in content (esp. text) and do not support easy further modification by users.
We benchmark Stable-Diffusion 2 \citep{rombach2022high} and DALL-E 3 \citep{betker2023improving} by asking them to output slides given the natural language instructions.
We adjust our reference-based evaluation procedure by first segmenting slide images into elements using Tesseract OCR\cite{smith2007overview} and further parse out the texts of the elements, then applying the default calculation process as in \S\ref{sec:eval-metrics}.
For the \textit{detailed instruction with images} setting, we also report the results of the end-to-end image generation methods, marked with a ``*'' to indicate that they do not actually use the image inputs.

\begin{table}
\centering
\vspace{-2mm}
\resizebox{0.47\textwidth}{!}{
\begin{tabular}{l|cccc|cccc} 
\toprule 
\multicolumn{1}{c|}{\multirow{2}{*}{Method}} & \multicolumn{4}{c|}{Detailed Instructions Only} & \multicolumn{4}{c}{High-Level Instructions} \\
\multicolumn{1}{c|}{} & {exec} & {ref-based} & {ref-free} & {overall} & {exec} & {ref-based} & {ref-free} & {overall} \\
\midrule 
\hlrowblue \multicolumn{9}{c}{\it End-to-End Image Generation} \\ 
\midrule 
{SD2} & {\bf 100.0} & {48.0} & {35.5} & {48.0} & {\bf 100.0} & {47.7} & {31.5} & {47.7} \\ 
{DALLE 3} & {\bf 100.0} & {50.2} & {57.5} & {50.2} & {\bf 100.0} & {50.7} & {53.6} & {52.2} \\ 
\midrule 
\midrule 
\hlrow \multicolumn{9}{c}{\it Code Generation w/o \lib} \\ 
\midrule 
{LLaVA} & {17.9} & {56.9} & {47.4} & {$~~$9.3} & {19.5} & {50.2} & {47.3} & {$~~$9.5} \\ 
{DeepseekCoder} & {$~~$2.6} & {59.6} & {37.5} & {$~~$1.3} & {22.6} & {57.6} & {43.0} & {11.4} \\
{CodeLLaMA} & - & - & - & - & {21.0} & {57.9} & {54.4} & {12.2} \\
{LLaMA} & {$~~$4.6} & {61.4} & {35.1}& {$~~$2.8} & {$~~$8.7} & {55.6} & {50.1} & {$~~$4.8}  \\ 
{GPT-4o} &  {50.3} & {\bf 66.8} & {50.0} & {28.7} & {70.8} & {\bf 60.3} & {57.0} & {39.7}   \\
\midrule 
\midrule 
\hlrow \multicolumn{9}{c}{\it Code Generation w/ \lib} \\ 
\midrule 
{LLaVA} & {17.4} & {58.2} & {33.8} & {$~~$8.0} & {25.1} & {50.1} & {36.7} & {10.9} \\ 
{DeepseekCoder} & {24.1} & {57.1} & {43.4} & {12.1} & {31.8} & {53.0} & {48.7} & {16.2} \\
{CodeLLaMA} & - & - & - & - & {35.9} & {56.6} & {53.4} & {20.3} \\
{LLaMA} & {60.5} & {61.7} & {56.6} & {37.4} & {76.9} & {56.8} & {58.3} & {43.7} \\ 
{GPT-4o} & {87.7} & {64.2} & {\bf 65.8} & {\bf 56.3} & {97.4} & {60.1} & {\bf 71.2} & {\bf 58.5}   \\ 
\midrule 
{\textbf{\model}} & {89.2} & {61.9} & {58.7} & {55.2} & {86.6} & {55.2} & {61.5} & {47.8} \\ 
\bottomrule 
\end{tabular}
}
\caption{Results under \textit{detailed instruction only} and \textit{high-level instructions} settings. We assign 100\% execution success rates for all end-to-end image generation methods because they do not generate programs and would not suffer from execution errors. }
\label{tab:results-others}
\end{table} 
\subsection{Quantitative Results and Analysis}
\label{sec:4.2:results}
\autoref{tab:results-sufficient} shows the result of \textit{detailed instruction with images} scenario and \autoref{tab:results-others} shows the result of  {\it detailed instruction only} and {\it high-level instructions} scenarios.

In the top row of \autoref{tab:results-sufficient}, we first measure the scores of the reference slides, which shows that the quality of the human-created slides is among the highest.

Compared to the scores achieved by \gpt, smaller open-source models such as \textsc{Llama 3.1} and \llava barely produce any working slides out of the box. Although the significant gaps of $49.9$--$55.0$ points exist in the \textit{detailed instruction with images} setting, this gap shrinks to $22.2$--$34.6$ when no visuals are provided a priori, in {\it detailed instruction only} and {\it high-level instructions} scenarios (\autoref{tab:results-others}). This demonstrates significant challenges in obtaining images in slides.
In contrast to the low performance of open-weight models out-of-the-box, \model's performance approaches that of \gpt.

\noindent {\bf End-to-End Image Generation} 
When no visuals are provided, end-to-end image-generation methods perform worse than the best code-generation approaches in both the reference-based and reference-free metrics, especially in generating accurate content.
These methods also often produce creative figures without aligning with the design principles of slides, indicating its poorer controllability.

\noindent {\bf Effect of \lib} \quad
\lib brings observable gains in \llama and \llava in all three scenarios by at most $34.0$ points; and similarly increases the strong \gpt performance across scenarios, especially when no images are provided. This suggests the benefits of generating more modular and concise programs for structured visual design.

\noindent {\bf VLM vs. LLM} \quad
When no helper functions are presented, the one VLM that we tested (\llava) outperforms its LLM counterpart \llama in all scenarios by $5.1$--$7.5$ points. However, \llava shows limited ability in using functions presented in context, as demonstrated by the large margin the library-augmented \llama has over \llava ($12.1$--$26.2$).\
All LLMs (\llama, \gpt) perform worse when the instructions become less specified (\textit{detailed instruction with images} $\rightarrow$ \textit{detailed instruction only} $\rightarrow$ \textit{high-level instructions}). Nonetheless, \lib can greatly mitigate this degradation due to the loss of input specificity, and help models produce better outcomes across all three scenarios.
\subsection{Qualitative Case Study}

We illustrate several models-produced slides in \autoref{fig:qualitative-examples}.
For end-to-end image generation methods, the design is more creative and often more attractive, but the text does not constitute meaningful words, or even the characters themselves are not valid.

On the other hand, code generation methods, especially weaker \llama and \llava models, suffer more from visual layout --- elements often overlap with each other or exceed the canvas, making it challenging for the audience to obtain all information clearly. 

In contrast, \model generates slides with appropriate layouts without undesirable element overlaps. In addition, they better follow the user instructions and are not overly creative like the image generation methods.

\subsection{Perceptual Evaluation}
\begin{table}[htbp]
\resizebox{0.47\textwidth}{!}{
\centering
\begin{tabular}{c|cc|cc}
\toprule
\multicolumn{1}{c|}{\multirow{2}{*}{\bf Model Pairs}} & \multicolumn{2}{c|}{Detailed+Images} & \multicolumn{2}{c}{Detailed Only} \\
\multicolumn{1}{c|}{} & {t-stat} & \multicolumn{1}{c|}{p-val} & {t-stat} & {p-val} \\ \midrule
(GPT-4o, \llama) & 13.206          &     0.000             &8.630 &0.000 \\
(\model, \llama)  &   13.180             & 0.000 &2.955 & 0.004\\
(GPT-4o, \model) & -0.445              & 0.657 & 8.203 &0.000\\
\bottomrule
\end{tabular}}
\caption{Paired t-test results comparing model performance across \textit{detailed instruction only} setting and \textit{detailed instruction with images} setting. \model and GPT-4o outperforms \llama with a statistically significant difference in both settings.
}
\label{tab:ttest-results}
\end{table}
We performed a qualitative evaluation on 10 randomly selected slides from each domain generated by GPT-4o, Llama-8B, and \model under the \textit{detailed instruction with images} and \textit{detailed instruction only} settings. We also add the ground-truth reference slide to evaluate the performance gap between current models and human slide creators. We shuffle these slides and ask the annotators to rank each slide from 1-5 based on how likely they would be to use the slide. For the \textit{detailed instruction with images} setting, we collect 25 responses in total, and for \textit{detailed instruction only}, we collect 16 responses in total. 
We provide more details of the evaluation process in \S\ref{app:user_eval}. 

\begin{figure}[t]
    \centering
\includegraphics[width=0.5\textwidth]{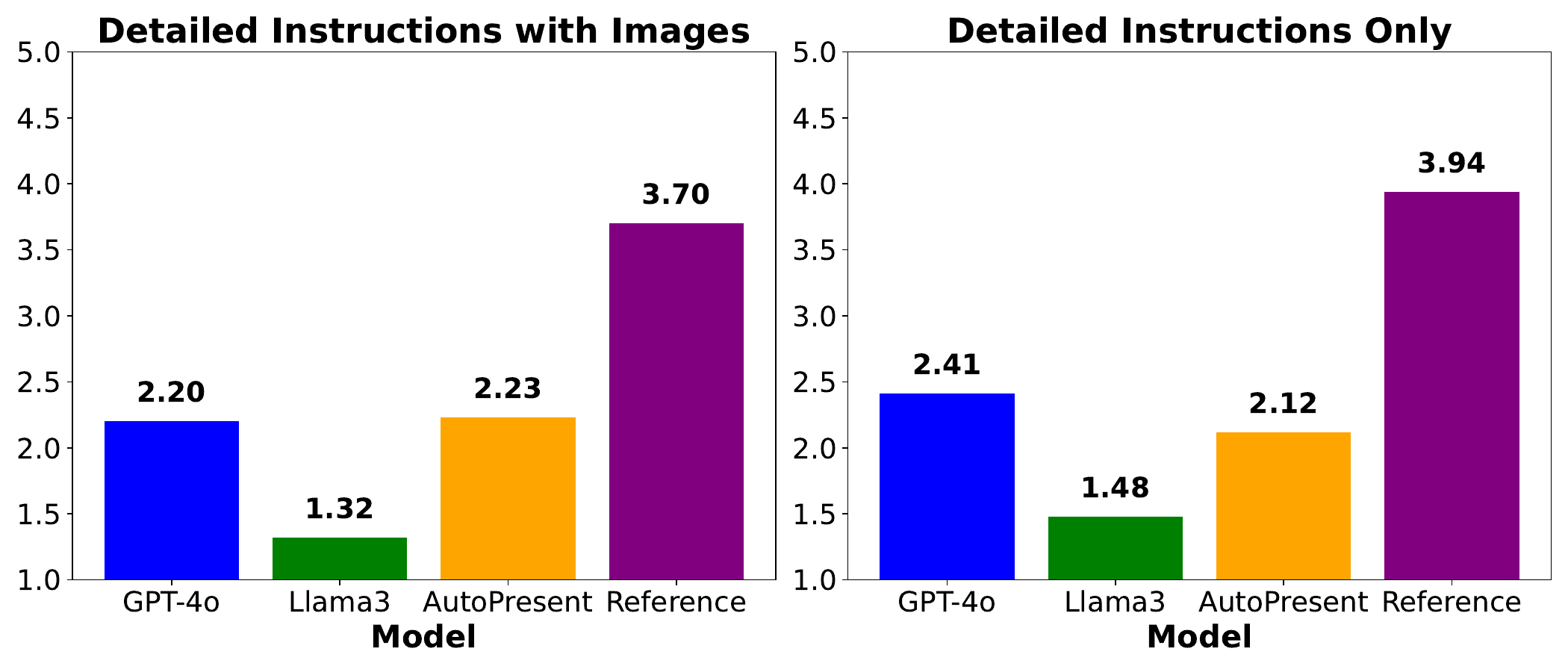}
    \caption{\textbf{Perceptual evaluation results on detailed instruction (1) with images and (2) only settings.} We ask the users to score the quality of each slide from 1-5 and report the average score of each model. The user reported preference on GPT-4o and \model compared with \llama, while still having a gap with human-designed slides. 
    }
    \label{fig:perceptual-eval}
\end{figure}

The result is shown in \autoref{fig:perceptual-eval}. 
By performing the paired $t$-test, we found differences between the models pairs in terms of user preferences, as shown in \autoref{tab:ttest-results}: 
(1) In both settings, \model and GPT-4o perform statistically significantly better than \llama. (2) In \textit{detailed instruction with images} setting, GPT-4o and \model has no significant differences (3) In the \textit{detailed instruction only} setting, \model is slightly worse than gpt-4o, aligning with our quantitative evaluations in \autoref{tab:results-others}.
All three models still have an overall performance gap compared with human-designed slides, indicating room for improvement on the slide generation task.
\subsection{Result after Iterative Refinement}

\begin{table}[ht!]
\centering
\vspace{-2mm}
\small
\resizebox{0.8\columnwidth}{!}{ 
\begin{tabular}{l|ccc} 
\toprule 
 Iteration & Detailed + Images & Detailed Only & High-Level\\ 
\midrule
{0} & {58.0} & {56.3} & {58.5}  \\ 
{1} &  {59.5} & {59.5} & {59.8}\\
{2} &  {59.3} & {\bf 60.1} & {61.3}\\
{3} &  {\bf 60.1} & {59.4} & {\bf 61.4}\\
\bottomrule 
\end{tabular}
}
\vspace{-2mm}
\caption{\textbf{Overall scores after applying multi-rounds of refinement in the three scenarios}, demonstrating that refinement boosts performance in all three scenarios.}
\label{tab:refinement}
\vspace{-1mm}
\end{table}

\begin{figure}[h!]
    \centering
\includegraphics[width=0.45\textwidth]{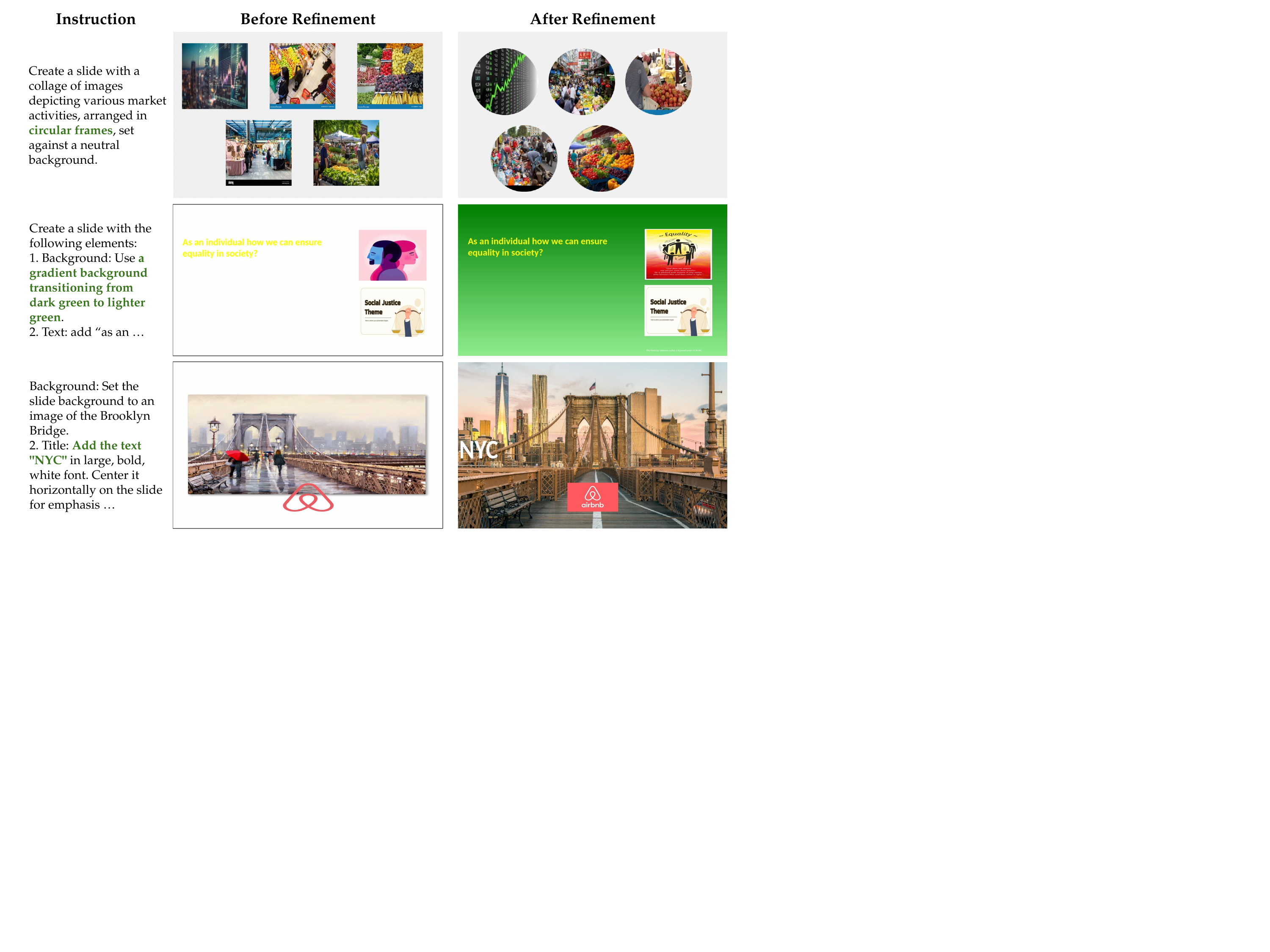}
    \caption{\textbf{Auto-refinement results with GPT-4o}, where the model further addresses some previously neglected instructions (marked in green), such as shape, background color, and text.}
    \label{fig:refinement}
\end{figure}

Finally, as shown by \autoref{tab:refinement}, we find that refinement improves model performance on all three challenges. By doing an ablation on the round of iterations, we find that while continued refinement often increases the scores, the first iteration usually gives the biggest performance improvement.
We present representative cases after doing one round of refinement in \autoref{fig:refinement}, which indicates that refinement can improve content layout and detailed controls on coloring and sizing.

\section{Related Work}
\label{related_work}

\noindent\textbf{Language and Vision Model-Based Agents} \quad
Agents based on large language models (LLMs) \citep{achiam2023gpt, dubey2024llama} and vision-language models (VLMs) \citep{alayrac2022flamingo, yang2024posterllava} have been widely adopted in various tasks such as web navigation \citep{webshop,zhou2024webarena,koh2024vwa}, software engineering \citep{yang2024swe,Yang2024SWEbenchMD}, and web development \citep{nakano2021webgpt,si2024design2code}.
Creation of presentation materials is another common task \citep{decktpus} that has both similarities and differences from these more widely examined tasks.

\noindent\textbf{Generating Programs for Vision Tasks} \quad
End-to-end image generation models such as diffusion \citep{ho2020denoising,rombach2022high,svgdreamer_xing_2023} and GAN \citep{goodfellow2014generative,sauer2023stylegan} are widely used at producing scenic images, yet falling short on more structured visuals such as websites and slides \citep{si2024design2code}.
Generating programs (i.e., image-editing actions) is a useful means to get structured visuals \citep{gupta2023visual, suris2023vipergpt, subramanian2023modular, wang2024trove, ge2025recursive}, 
including Tikz figures \citep{Belouadi2023AutomaTikZTS, belouadi2024detikzify}, SVG \citep{rodriguez2023starvector, sharma2024vision}, posters \citep{yang2024posterllava}, and user interfaces \citep{si2024design2code, peng2025dreamstruct}. 
However, they often require detailed inputs and are limited to specific, simple figure types, so they are still far from creating complex, editable presentation slides from scratch. Our work extends this line of research by formulating and benchmarking the natural-language-to-slide generation task.

\noindent\textbf{Automatic Slide Generation} \quad
Previous works on slide creation mostly focus on basic extraction from provided documents \citep{kan2007slideseer, hu2014ppsgen, sefid2019automatic, sun2021d2s, fu2022doc2ppt} or having models generate content given a topic \citep{al2005auto, winters2019automatically, gamma} without addressing how to organize content visually.
More recently, some benchmarks \citep{guo2023pptc, zhang2024pptcr} and methods \citep{gandhi2023natural} have emerged that follow detailed instructions for slide editing (e.g., adjust the font size of the title from 20 to 24) of an existing slide. 
In contrast, we synthesize more complex and structured programs that can generate slides from scratch, including content creation, visual arrangement, and fine-grained editing, instead of refining an existing slide.
\section{Conclusion and Limitations}
In this work, we address the challenge of creating structured visuals from scratch.
Specifically, we introduced \data, the first benchmark for automatic slide generation with evaluation metrics based on and free of reference slides. 
We benchmark multiple end-to-end image and program generation approaches, and demonstrate that \model with \lib achieves comparable performance with the top \gpt model.
Our further exploration in iterative refinement also reveals certain effectiveness in self-refinement.
This work is an initial step towards automated generation of structured visuals. Specifically, it focuses on single-slide generation and produces full slide code in a single pass, without leveraging iterative design workflows. Future research could address these limitations by expanding to full slide decks, adopting gradual and interactive slide generation, and incorporating slide-specific features like animations. Further, integrating more design principles, such as optimizing for attention capture and information clarity, would be crucial for making generated slides more impactful and effective.

\section*{Acknowledgment}
We would like to thank Yutong Bai for helping us draft \autoref{fig:teaser} and providing feedback on the paper, David Chan for providing detailed suggestions on the introduction, and Frank Xu and Sean Welleck for discussions at the initial stage of this project. Junyi Zhang and Haven Feng for feedback on the project.

{
    \small
    \bibliographystyle{ieeenat_fullname}
    \bibliography{main}
}

\clearpage
\appendix
\section{\data Details}
\label{app:implementation}
\subsection{Slide Deck Domains}
The 10 domains we cover in \data include: 
\begin{enumerate}
\small
    \item Art Photos
    \item Business
    \item Career
    \item Design
    \item Entrepreneur
    \item Environment
    \item Food
    \item Marketing
    \item Social Media
    \item Technology
\end{enumerate}
\subsection{Slide Deck Source} There existis large amount of slide decks on the internet including Google Search, Bing Search etc. For convenience, we collect a list of slides from the \url{slideshare.com} website.
\subsection{Slide Deck Statistics Per Domain} 
\begin{figure}[htbp]
\centering
\includegraphics[width=\linewidth]{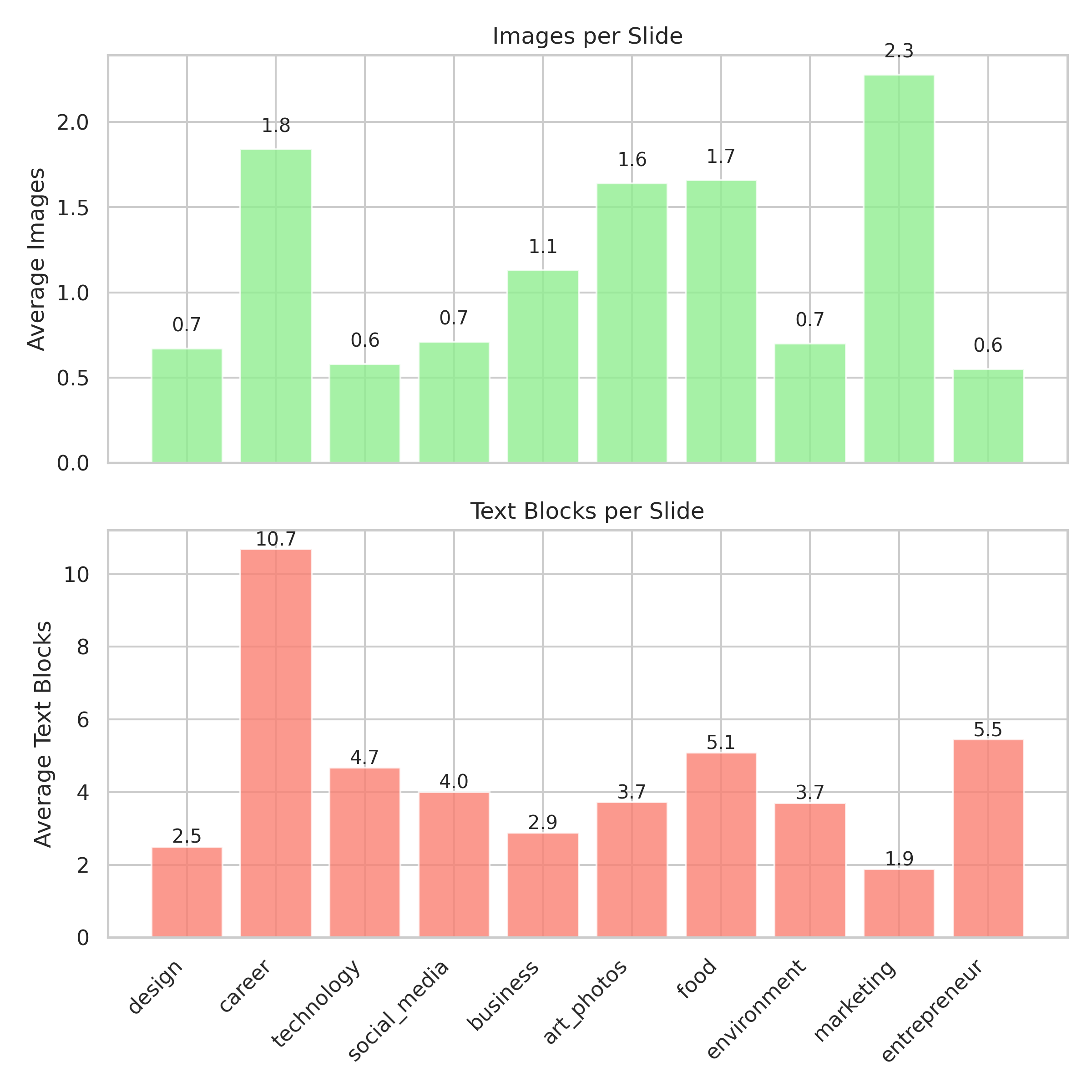}
\caption{SlidesBench statistics on different domains.}
\label{fig:stats_}
\end{figure}

The \textit{average images per domain} and \textit{average text blocks per domain} are shown in \autoref{fig:stats_}.

\section{\lib Details}
\label{app:library}
In this section, we provide the detailed documentation and examples for all functions in our \lib. 

\subsection{\lib Implementation}
\autoref{list:basic-apis} shows the basic functions and \autoref{list:image-apis} shows the image-oriented functions.
\begin{figure*}[htbp]
\lstset{
    language=Python,
    basicstyle=\footnotesize\ttfamily,
    keywordstyle=\color{blue},
    stringstyle=\color{red},
    commentstyle=\color{gray},
    breaklines=true,
}
\begin{lstlisting}
add_title(slide, text, font_size, font_color, background_color)
"""Add a title text to the slide with custom font size and font color (RGB tuple).
Args:
    slide: Slide object as in pptx library
    text: str, Title text to be added
    font_size: int, Font size in int (point size), e.g., 44
    font_color: tuple(int,int,int), RGB color, e.g., (0, 0, 0)
    background_color: Optional, tuple(int,int,int), RGB color, e.g., (255, 255, 255)
Rets:
    slide: Slide object with the title added
"""

add_text(slide, text, coords, font_size, bold, color, background_color, auto_size)
"""Add a text box at a specified location with custom text and color settings.
Args:
    slide: Slide object as in pptx library
    text: str, Text to be added
    coords: list(float), [left, top, width, height] in inches
    font_size: int, Font size in int (point size), e.g., 20
    bold: bool, True if bold-type the text, False otherwise
    color: tuple(int,int,int), RGB color, e.g., (0, 0, 0)
    background_color: Optional, tuple(int,int,int), RGB color, e.g., (255, 255, 255)
    auto_size: bool, True if auto-size the text box, False otherwise
Rets:
    slide: Slide object with the text box added
"""

add_bullet_points(slide, bullet_points, coords, font_size, color, background_color)
"""Add a text box with bullet points.
Args:
    slide: Slide object as in pptx library
    bullet_points: list(str), List of texts to be added as bullet points
    coords: list(float), [left, top, width, height] in inches
    font_size: int, Font size in int (point size), e.g., 18
    color: tuple(int,int,int), RGB color, e.g., (0, 0, 0)
    background_color: Optional, tuple(int,int,int), RGB color, e.g., (255, 255, 255)
Rets:
    slide: Slide object with the bullet points added
"""

add_image(slide, image_path, coords)
"""Add an image in the provided path to the specified coords and sizes.
Args:
    slide: Slide object as in pptx library
    image_path: str, Path to the image file
    coords: list(float), [left, top, width, height] in inches
Rets:
    slide: Slide object with the image added
"""

set_background_color(slide, color)
"""Set background color for the current slide.
Args:
    slide: Slide object as in pptx library
    color: tuple(int,int,int), RGB color, e.g., (255, 255, 255)
Rets:
    modified slide object
"""
\end{lstlisting}
\vspace{-4mm}
\caption{Documentation for the basic functions in our \lib.}
\label{list:basic-apis}
\end{figure*}

\begin{figure*}[htbp]
\lstset{
    language=Python,
    basicstyle=\footnotesize\ttfamily,
    keywordstyle=\color{blue},
    stringstyle=\color{red},
    commentstyle=\color{gray},
    breaklines=true,
}
\begin{lstlisting}
google_search_screenshot(question, save_path)
"""Search a question on Google, and take a screenshot of the search result.
Save the screenshot to save_path, and return the path.
Args:
    question: str, The question to search on Google.
    save_path: str, The path to save the screenshot.
Returns:
    The path of the saved screenshot.
"""

search_image(query, save_path)
"""Search for an image on Google and download the result to save_path.
Args:
    query: str, The query to search for.
    save_path: str, The path to save the downloaded image.
Rets:
    the save_path.
"""

generate_image(query, save_path)
"""Generate an image using diffusion model based on a text query, and save the image to the path.
Args:
    query: str, The text query to generate the image.
    save_path: str, The path to save the generated image.
Rets:
    The path of the saved image
"""
\end{lstlisting}
\vspace{-4mm}
\caption{Documentation for the image-oriented functions in our \lib.}
\label{list:image-apis}
\end{figure*}

\subsection{\lib Usage Example}
\label{app:lib:example-programs}

\autoref{list:lib-usage} shows two example programs using multiple \lib functions to produce slides.

\begin{figure*}[htbp]
\lstset{
    language=Python,
    basicstyle=\footnotesize\ttfamily,
    keywordstyle=\color{blue},
    stringstyle=\color{red},
    commentstyle=\color{gray},
    breaklines=true,
}
\begin{lstlisting}
# Create slide with the title 'NLP Can Answer Questions' in large, bolded font in the top center of the page. Below it, put a screenshot of the google search result of the question 'Where was the first movie theater in the U.S?' in the middle of the page.

from pptx import Presentation
from pptx.util import Inches, Pt
from library import add_text, google_search_screenshot, add_image

presentation = Presentation()
presentation.slide_width = Inches(16)
presentation.slide_height = Inches(9)

slide_layout = presentation.slide_layouts[0] # choose a layout template
slide = presentation.slides.add_slide(slide_layout)
add_text(slide, "NLP Can Answer Questions", coords=(1, 0.5, 8, 1), font_size=36)
img_path = google_search_screenshot("Where was the first movie theater in the U.S?", save_path="screenshot.png")
add_image(slide, "screenshot.png", coords=(2.5, 2, 6, 4))
presentation.save("target_path.pptx")


# Create a slide titled 'Interior Design' in bold, dark-green color in the center of the page. For the background, consider using a picture with a color, artistic vibe, ensure enough contrast between the colors of text and background.

from pptx import Presentation
from pptx.util import Inches, Pt
from library import generate_image, add_image, add_text

presentation = Presentation()
presentation.slide_width = Inches(16)
presentation.slide_height = Inches(9)
slide_layout = presentation.slide_layouts[5] # choose a layout template
slide = presentation.slides.add_slide(slide_layout)

background_img = generate_image("An colorful, artistic background", "colorful.png")
add_image(slide, "colorful.png", coords=(0.0, 0.0, 16, 9))
add_text(slide, 'Interior Design', coords=(0.0, 2.4, 13.3, 1.3), font_size=80, bold=True, color=(0, 0, 0), background_color=(255, 255, 255), auto_size=True)
presentation.save("path.pptx")
```
\end{lstlisting}
\vspace{-4mm}
\caption{Example programs to produce slides using \lib.}
\label{list:lib-usage}
\end{figure*}
\begin{figure}[htbp]
\centering
\includegraphics[width=\linewidth]{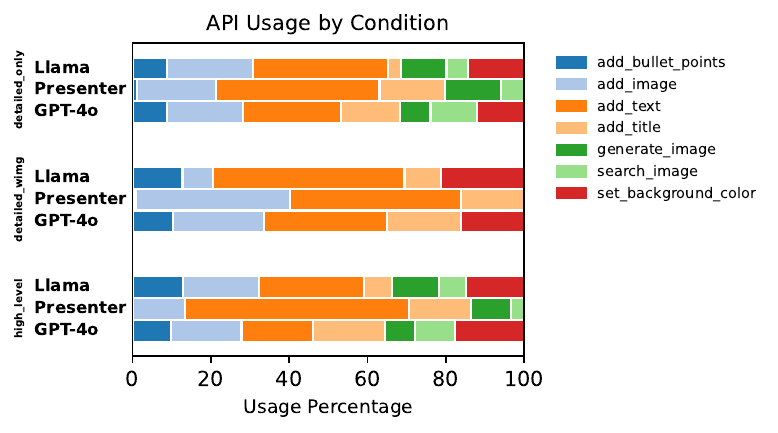}
\caption{The percentage of each action taken by different models.}
\label{fig:actions_}
\end{figure}

\subsection{\lib Usage Percentage}
The \textit{percentage of each action} taken by GPT-4o, \model, and Llama-3.1 in all 3 scenarios are reported in \autoref{fig:actions_}. On average, the most common actions are \texttt{add\_text} (36.3\%), \texttt{add\_image} (20.3\%), and \texttt{add\_title} (13.5\%).

\section{Training Details for \model}
\label{app:training}

The training parameters for \model are summarized in Table~\ref{tab:training-details}.

\begin{table}[htbp]
\small
\centering
\begin{tabular}{c|c}
\hline
\textbf{Parameter}                & \textbf{Value}                            \\ \hline
\multicolumn{2}{c}{\textbf{LoRA Parameters}}                               \\ \hline
LoRA rank                         & 128                                      \\ \hline
LoRA alpha                        & 32                                       \\ \hline
LoRA dropout                      & 0                                        \\ \hline
Random state                      & 3407                                     \\ \hline
RS-LoRA                           & Disabled                                 \\ \hline
LoFT-Q config                     & None                                     \\ \hline
\multicolumn{2}{c}{\textbf{Trainer Parameters}}                            \\ \hline
Batch size (per device)           & 1                                        \\ \hline
Gradient accumulation steps       & 2                                        \\ \hline
Warmup steps                      & 20                                       \\ \hline
Epochs                            & 1                                        \\ \hline
Learning rate                     & 3e-4                                     \\ \hline
Mixed precision                   & FP16                                     \\ \hline
Weight decay                      & 0.01                                     \\ \hline
Scheduler                         & Linear                                   \\ \hline
Seed                              & 3407                                     \\ \hline
\end{tabular}
\caption{Training details for \model. LoRA and Trainer parameters are described in detail.}
\label{tab:training-details}
\end{table}

\section{Refinement Details}
\label{app:refinement}
We provide the prompts that we used for auto-refinement in \autoref{list:refinement-prompt}. We input the instruction and the in-context examples, the previous code generated by the model, and the snapshot of the slide generated by executing this code to the model and let it correct the code.

\begin{figure*}[htbp]
\lstset{
    language=Python,
    basicstyle=\footnotesize\ttfamily,
    keywordstyle=\color{blue},
    stringstyle=\color{red},
    commentstyle=\color{gray},
    breaklines=true,
}
\begin{lstlisting}
"""
You are an expert presentation slides designer who creates modern, fashionable, and stylish slides using Python code. Your job is to generate the required PPTX slide by writing and executing a Python script. Make sure to follow the guidelines below and do not skip any of them:
1.  Ensure your code can successfully execute. If needed, you can also write tests to verify your code.
2. Maintain proper spacing and arrangements of elements in the slide: make sure to keep sufficient spacing between different elements; do not make elements overlap or overflow to the slide page.
3. Carefully select the colors of text, shapes, and backgrounds, to ensure all contents are readable.
4. The slides should not look empty or incomplete. When filling the content in the slides, maintain good design and layout.

Follow the instruction below to create the slide. 
If the instruction is long and specific, follow the instruction carefully and add all elements as required; 
if it is short and concise, you will need to create some content (text, image, layout) and implement it into the slide.
If you need to use the provided images, refer to the image file names in the instructions.

Finally, your code should save the pptx file to path "output.pptx"

API Libraries:
# INSERT_API_DESCRIPTIONS_HERE

## Examples
# INSERT_IN_CONTEXT_EXAMPLES_HERE

Modification Task:
Instruction: INSERT_INSTRUCTION_HERE

Previous Code:
INSERT_PREV_CODE_HERE

Slide Snapshot : See image.
Task: Based on the observed drawbacks in the provided slide image, modify the existing code accordingly to improve the slide's design and functionality.

Your modification:
def generate_presentation():
"""

\end{lstlisting}
\vspace{-4mm}
\caption{\textbf{Prompt we used for Auto-Refinement.} The model receives the APIs and instruction, the previous generated slide and code, and is tasked to re-write the code to do slide refinement.}
\label{list:refinement-prompt}
\end{figure*}

\section{Detailed Results}
\label{app:results}
We report two sets of evaluation metrics (reference-based and reference-free) in both their average value on all slides (i.e., un-weighted by execution success) and on successfully rendered slides (i.e., weighted by execution success).

\subsection{Detailed Instructions with Images}

\autoref{tab:results-sufficient-weighted} shows all metrics down-weighted by the execution success rate; \autoref{tab:results-sufficient-unweighted} shows reference-based and reference-free metrics without down-weighting by execution success.

\begin{table*}[htbp]
\centering
\footnotesize
\vspace{-2mm}
\begin{tabular}{l|c|cccc|cccc|c} 
\toprule 
\multicolumn{1}{c|}{\multirow{2}{*}{\bf Method}} & \multirow{2}{*}{\bf Execution\%} & \multicolumn{4}{c|}{\bf Reference-Based} & \multicolumn{4}{c|}{\bf Reference-Free}  & \multirow{2}{*}{\bf Average} \\ 
\multicolumn{1}{c|}{} & {} & {\bf block} & {\bf text} & {\bf color} & {\bf position} &{\bf text} & {\bf image}& {\bf layout} & {\bf color} & {} \\
\midrule
{Human} & {100.0} & \multicolumn{4}{c|}{-}  & {59.7} & {81.5} & {73.5} & {65.7} & - \\ 
\midrule 
\hlrow \multicolumn{11}{c}{\it Code Generation w/o Library} \\ 
\midrule 
{LLaVA (7B)} & {11.3} & {$~~$7.0} & {11.0} & {$~~$0.7} & {$~~$8.0} & {$~~$4.7} & {11.3} & {$~~$3.3} & {$~~$2.9} & {$~~$6.1} \\ 
{LLaMA (8B)} & {$~~$2.1} & {$~~$1.5} & {1.9} & {$~~$0.3} & {$~~$1.7} & {$~~$1.0} & {$~~$0.2} & {$~~$1.0} & {$~~$1.0} & {$~~$1.3} \\
{GPT-4o} & {\bf 89.2} & {74.3} & {\bf 80.7} & {$~~$9.4} & {\bf 68.7} & {46.3} & {64.9} & {47.9} & {48.8} & {55.1} \\ 
\midrule
{\textbf{\model} (ours)} & {79.0} & {53.5} & {63.0} & {$~~$8.6} & {60.0} & {35.8} & {49.5} & {42.8} & {48.1} & {46.3} \\ 
\midrule
\hlrow \multicolumn{11}{c}{\it Code Generation w/ Expert-Designed Library} \\ 
\midrule 
{LLaVA (7B)} & {20.0} & {16.1} & {16.1} & {$~~$0.7} & {12.8} & {$~~$7.5} & {$~~$9.6} & {$~~$5.9} & {$~~$8.7} & {$~~$9.7} \\ 
{LLaMA (8B)} & {54.4} & {42.6} & {49.6} & {$~~$4.1} & {37.8} & {25.0} & {37.1} & {25.9} & {28.9} & {33.5} \\ 
{GPT-4o} & {86.7} & {74.7} & {80.2} & {11.0} & {66.1} & {\bf 47.3} & {\bf 72.5} & {\bf 61.1} & {51.4} & {\bf 58.0} \\ 
\midrule
{\textbf{\model} (ours)} & {84.1} & {70.8} & {77.5} & {\bf 15.2} & {56.5} & {40.2} & {61.6} & {49.3} & {\bf 54.4} & {55.0} \\ 
\bottomrule 
\end{tabular}
\vspace{-1mm}
\caption{Slide generation results (weighted by execution success) under the {\it detailed instructions with images} scenario.}
\label{tab:results-sufficient-weighted}
\vspace{1mm}
\end{table*}

\begin{table*}[htbp]
\centering
\footnotesize
\vspace{-2mm}
\begin{tabular}{l|c|cccc|cccc|c} 
\toprule 
\multicolumn{1}{c|}{\multirow{2}{*}{\bf Method}} & \multirow{2}{*}{\bf Execution\%} & \multicolumn{4}{c|}{\bf Reference-Based} & \multicolumn{4}{c|}{\bf Reference-Free}  & \multirow{2}{*}{\bf Avg} \\ 
\multicolumn{1}{c|}{} & {} & {\bf block} & {\bf text} & {\bf color} & {\bf pos} &{\bf text} & {\bf img}& {\bf layout} & {\bf color} & {} \\
\midrule
{Human} & {100.0} & \multicolumn{4}{c|}{-} & {59.7} & {81.5} & {73.5} & {65.7} & - \\ 
\midrule 
\hlrow \multicolumn{11}{c}{\it Code Generation w/o Library} \\ 
\midrule 
{LLaVA (7B)} & {11.3} & {61.9} & {\bf 97.3} & {$~~$6.2} & {70.8} & {41.6} & {\bf 100.0} & {29.2} & {25.7} & {$~~$6.1} \\ 
{LLaMA (8B)} & {$~~$2.1} & {74.0} & {94.6} & {12.5} & {\bf 81.2} & {50.0} & {$~~$8.3} & {50.0} & {50.0} & {$~~$1.3} \\
{GPT-4o} & {\bf 89.2} & {83.3} & {91.6} & {10.5} & {77.0} & {51.9} & {72.8} & {53.7} & {54.7} & {55.1} \\ 
\midrule
{\textbf{\model}} & {79.0} & {67.7} & {79.7} & {10.9} & {75.9} & {45.3} & {$~~$62.7} & {54.2} & {60.9} & {46.3} \\ 
\midrule
\hlrow \multicolumn{11}{c}{\it Code Generation w/ Expert-Designed Library} \\ 
\midrule 
{LLaVA (7B)} & {20.0} & {80.5} & {80.5} & {$~~$3.5} & {64.0} & {37.5} & {48.0} & {29.5} & {43.5} & {$~~$9.7} \\ 
{LLaMA (8B)} & {54.4} & {78.3} & {91.2} & {7.5} & {69.5} & {46.0} & {68.2} & {47.6} & {53.1} & {33.5} \\ 
{GPT-4o} & {86.7} & {\bf 86.2} & {92.5} & {12.7} & {76.3} & {\bf 54.6} & {83.7} & {\bf 70.5} & {59.4} & {\bf 58.0} \\ 
\midrule
{\textbf{\model} (ours)} & {84.1} & {84.2} & {92.2} & {\bf 18.1} & {67.2} & {47.8} & {73.2} & {58.6} & {\bf 64.7} & {55.0} \\ 
\bottomrule 
\end{tabular}
\vspace{-2mm}
\caption{Slide generation results (un-weighted by execution success) under the {\it detailed instructions with images} scenario.}
\label{tab:results-sufficient-unweighted}
\end{table*}


\subsection{Detailed Instructions Only}
\autoref{tab:results-noimage-weighted} shows all metrics down-weighted by the execution success rate; \autoref{tab:results-noimage-unweighted} shows reference-based and reference-free metrics without down-weighting by execution success.

\begin{table*}[htbp]
\centering
\footnotesize
\vspace{-2mm}
\begin{tabular}{l|c|cccc|cccc|c} 
\toprule 
\multicolumn{1}{c|}{\multirow{2}{*}{\bf Method}} & \multicolumn{1}{c|}{\multirow{2}{*}{\bf Execution\%}} & \multicolumn{4}{c|}{\bf Reference-Based} & \multicolumn{4}{c|}{\bf Reference-Free}  & \multirow{2}{*}{\bf Average} \\ 
\multicolumn{1}{c|}{} & {} & {\bf block} & {\bf text} & {\bf color} & {\bf position} &{\bf text} & {\bf image}& {\bf layout} & {\bf color} & {} \\
\midrule 
\hlrowblue \multicolumn{11}{c}{\it End-to-End Image Generation} \\ 
\midrule 
{Stable-Diffusion} & {\bf 100.0} & {74.5} & {33.4} & {$~~$9.0} & {75.0} & {19.6} & {45.1} & {36.9} & {40.5} & {48.0} \\ 
{DALLE 3} & {\bf 100.0} & {\bf 75.5} & {39.9} & {$~~$9.2} & {76.1} & {32.7} & {\bf 87.3} & {56.7} & {53.4} & {50.2} \\ 
\midrule 
\midrule 
\hlrow \multicolumn{11}{c}{\it Code Generation w/o Library} \\ 
\midrule 
{LLaVA (7B)} & {17.9} & {12.2} & {16.3} & {$~~$1.4} & {12.4} & {$~~$7.9} & {15.3} & {$~~$5.7} & {$~~$5.0} & {$~~$9.5} \\ 
{LLaMA (8B)} & {4.6} & {63.0} & {87.0} & {\bf 17.4} & {\bf 80.4} & {30.4} & {19.6} & {41.3} & {47.8} & {$~~$2.8}  \\ 
{GPT-4o} &  {50.3} & {42.2} & {50.0}  & {$~~$6.0} & {39.8} & {27.1} & {15.3} & {29.0} & {29.2} & {32.2}   \\
\midrule 
\midrule 
\hlrow \multicolumn{11}{c}{\it Code Generation w/ Expert-Designed Library} \\ 
\midrule 
{LLaVA (7B)} & {17.4} & {15.6} & {15.5} & {$~~$0.9} & {10.5} & {$~~$5.7} & {$~~$6.2} & {$~~$4.1} & {$~~$7.5} & {$~~$8.3} \\ 
{LLaMA (8B)} & {60.5} & {45.1} & {55.5} & {$~~$5.2} & {43.6} & {29.5} & {44.3} & {29.6} & {33.4} & {37.4} \\ 
{GPT-4o} & {87.7} & {72.3} & {80.8} & {$~~$6.0} & {65.9} & {\bf 46.6} & {73.0} & {\bf 58.5} & {52.9} & {\bf 56.3}   \\ 
\midrule 
{\textbf{\model} (ours)} & {89.2} & {70.2} & {\bf 82.7} & {$~~$9.3} & {58.5} & {43.0} & {47.7} & {55.3} & {\bf 63.2} & {55.2} \\
\bottomrule 
\end{tabular}
\caption{Results (weighted by execution success) under \textit{detailed instructions only} scenario.}
\label{tab:results-noimage-weighted}
\end{table*}

\begin{table*}[htbp]
\centering
\footnotesize
\vspace{-2mm}
\begin{tabular}{l|c|cccc|cccc|c} 
\toprule 
\multicolumn{1}{c|}{\multirow{2}{*}{\bf Method}} & \multicolumn{1}{c|}{\multirow{2}{*}{\bf Execution\%}} & \multicolumn{4}{c|}{\bf Reference-Based} & \multicolumn{4}{c|}{\bf Reference-Free}  & \multirow{2}{*}{\bf Overall} \\ 
\multicolumn{1}{c|}{} & {} & {\bf block} & {\bf text} & {\bf color} & {\bf position} &{\bf text} & {\bf image}& {\bf layout} & {\bf color} & {} \\
\midrule 
\hlrowblue \multicolumn{11}{c}{\it End-to-End Image Generation} \\ 
\midrule 
{Stable-Diffusion} & {\bf 100.0} & {74.5} & {33.4} & {$~~$9.0} & {75.0} & {19.6} & {45.1} & {36.9} & {40.5} & {48.0} \\ 
{DALLE 3} & {\bf 100.0} & {75.5} & {39.9} & {$~~$9.2} & {76.1} & {32.7} & {\bf 87.3} & {56.7} & {53.4} & {50.2} \\ 
\midrule 
\midrule 
\hlrow \multicolumn{11}{c}{\it Code Generation w/o Library} \\ 
\midrule 
{LLaVA (7B)} & {17.9} & {68.2} & {91.1} & {$~~$7.8} & {69.3} & {44.1} & {85.8} & {31.8} & {27.9} & {9.5} \\ 
{LLaMA (8B)} & {4.6} & {2.9} & {$~~$4.0} & {$~~$0.8} & {$~~$3.7} & {$~~$1.4} & {$~~$0.9} & {$~~$1.9} & {$~~$2.2} & {$~~$2.8}  \\ 
{GPT-4o} &  {50.3} & {83.9} & {92.4}  & {\bf 11.9} & {\bf 79.1} & {\bf 53.9} & {30.4} & {57.7} & {58.1} & {32.2}   \\
\midrule 
\midrule 
\hlrow \multicolumn{11}{c}{\it Code Generation w/ Expert-Designed Library} \\ 
\midrule 
{LLaVA (7B)} & {17.4} & {\bf 89.7} & {89.1} & {$~~$5.2} & {60.3} & {32.8} & {35.6} & {23.6} & {43.1} & {$~~$8.3} \\ 
{LLaMA (8B)} & {60.5} & {74.5} & {91.7} & {$~~$8.6} & {72.1} & {48.8} & {73.2} & {29.6} & {48.9} & {37.4} \\ 
{GPT-4o} & {87.7} & {82.4} & {92.2} & {$~~$6.9} & {75.2} & {53.1} & {83.3} & {\bf 66.7} & {60.3} & {\bf 56.3}   \\ 
\midrule 
{\textbf{\model} (ours)} & {89.2} & {78.7} & {\bf 92.7} & {10.4} & {65.6} & {48.2} & {53.5} & {62.0} & {70.9} & {55.2} \\
\bottomrule 
\end{tabular}
\vspace{-1mm}
\caption{Results (un-weighted by execution success) under \textit{detailed instructions only} scenario.}
\label{tab:results-noimage-unweighted}
\vspace{1mm}
\end{table*}

\subsection{High-Level Instructions Challenge}
\autoref{tab:results-hl-weighted} shows all metrics down-weighted by the execution success rate; \autoref{tab:results-hl-unweighted} shows reference-based and reference-free metrics without down-weighting by execution success.

\begin{table*}[htbp]
\centering
\footnotesize
\vspace{-2mm}
\begin{tabular}{l|c|cccc|cccc|c} 
\toprule 
\multicolumn{1}{c|}{\multirow{2}{*}{\bf Method}} & \multicolumn{1}{c|}{\multirow{2}{*}{\bf Execution\%}} & \multicolumn{4}{c|}{\bf Reference-Based} & \multicolumn{4}{c|}{\bf Reference-Free}  & \multirow{2}{*}{\bf Average} \\ 
\multicolumn{1}{c|}{} & {} & {\bf block} & {\bf text} & {\bf color} & {\bf position} &{\bf text} & {\bf image}& {\bf layout} & {\bf color} & {} \\
\midrule 
\hlrowblue \multicolumn{11}{c}{\it End-to-End Image Generation} \\ 
\midrule 
{Stable-Diffusion} & {\bf 100.0} & {72.0} & {33.2} & {$~~$8.3} & {77.2} & {$~~$3.3} & {49.3} & {35.6} & {37.8} & {47.7} \\
{DALLE 3} & {\bf 100.0} & {73.5} & {48.2} & {$~~$7.6} & {\bf 77.3} & {14.9} & {\bf 89.7} & {57.2} & {52.4} & {51.7} \\
\midrule 
\midrule 
\hlrow \multicolumn{11}{c}{\it CodeGen-based Methods w/o Library} \\ 
\midrule 
{LLaVA (7B)} & {19.5} & {14.9} & {13.2} & {$~~$1.7} & {13.6} & {$~~$8.0} & {16.8} & {$~~$5.9} & {$~~$6.2} & {10.0} \\ 
{LLaMA (8B)} & {8.7} & {7.6} & {$~~$6.3} & {$~~$0.7} & {$~~$4.7} & {$~~$4.6} & {$~~$2.4} & {$~~$5.0} & {$~~$5.4} & {$~~$4.8} \\ 
{GPT-4o} &  {70.8} & {54.6} & {54.2} & {$~~$7.5} & {54.4} & {42.4} & {19.2} & {51.9} & {48.0} & {39.0} \\ 
\midrule 
\midrule 
\hlrow \multicolumn{11}{c}{\it CodeGen-based Methods w/ Library} \\ 
\midrule 
{LLaVA (7B)} & {25.1} & {20.4} & {17.8} & {$~~$1.6} & {15.4} & {$~~$9.2} & {$~~$9.7} & {$~~$6.9} & {11.0} & {11.5} \\ 
{LLaMA (8B)} & {76.9} & {55.4} & {58.3} & {$~~$5.6} & {55.7} & {39.5} & {56.5} & {40.3} & {43.0} & {43.7} \\ 
{GPT-4o} & {97.4} & {\bf 77.0} & {\bf 75.8} & {$~~$7.7} & {73.7} & {\bf 59.7} & {73.8} & {\bf 78.7} & {65.4} & {\bf 58.5} \\
\midrule 
{\textbf{\model} (ours)} & {86.6} & {63.5} & {66.4} & {10.2} & {51.1} & {41.4} & {34.2} & {64.0} & {\bf 73.3} & {47.8} \\ 
\bottomrule 
\end{tabular}
\vspace{-1mm}
\caption{Results (weighted by execution success) under \textit{high-level instructions} scenario.}
\label{tab:results-hl-weighted}
\vspace{1mm}
\end{table*}

\begin{table*}[htbp]
\centering
\footnotesize
\vspace{-2mm}
\begin{tabular}{l|c|cccc|cccc|c} 
\toprule 
\multicolumn{1}{c|}{\multirow{2}{*}{\bf Method}} & \multicolumn{1}{c|}{\multirow{2}{*}{\bf Execution\%}} & \multicolumn{4}{c|}{\bf Reference-Based} & \multicolumn{4}{c|}{\bf Reference-Free}  & \multirow{2}{*}{\bf Average} \\ 
\multicolumn{1}{c|}{} & {} & {\bf block} & {\bf text} & {\bf color} & {\bf position} &{\bf text} & {\bf image}& {\bf layout} & {\bf color} & {} \\
\midrule 
\hlrowblue \multicolumn{11}{c}{\it End-to-End Image Generation} \\ 
\midrule 
{Stable-Diffusion} & {\bf 100.0} & {72.0} & {33.2} & {$~~$8.3} & {77.2} & {$~~$3.3} & {49.3} & {35.6} & {37.8} & {47.7} \\
{DALLE 3} & {\bf 100.0} & {73.5} & {48.2} & {$~~$7.6} & {\bf 77.3} & {14.9} & {\bf 89.7} & {57.2} & {52.4} & {51.7} \\
\midrule 
\midrule 
\hlrow \multicolumn{11}{c}{\it CodeGen-based Methods w/o Library} \\ 
\midrule 
{LLaVA (7B)} & {19.5} & {76.4} & {67.7} & {$~~$8.7} & {69.7} & {41.0} & {86.2} & {30.3} & {31.8} & {10.0} \\ 
{LLaMA (8B)} & {$~~$8.7} & {\bf 87.4} & {72.4} & {$~~$8.0} & {54.0} & {52.9} & {27.6} & {57.5} & {62.1} & {$~~$4.8} \\ 
{GPT-4o} &  {70.8} & {77.1} & {76.8} & {10.6} & {76.8} & {59.9} & {27.1} & {73.3} & {67.8} & {39.0} \\ 
\midrule 
\midrule 
\hlrow \multicolumn{11}{c}{\it CodeGen-based Methods w/ Library} \\ 
\midrule 
{LLaVA (7B)} & {25.1} & {81.3} & {70.9} & {$~~$6.4} & {61.4} & {36.7} & {38.6} & {27.5} & {43.8} & {11.5} \\ 
{LLaMA (8B)} & {76.9} & {72.0} & {75.7} & {$~~$7.3} & {72.4} & {51.3} & {73.4} & {52.4} & {55.9} & {43.7} \\ 
{GPT-4o} & {97.4} & {79.0} & {\bf 77.8} & {$~~$7.9} & {75.6} & {\bf 61.3} & {75.8} & {\bf 80.7} & {67.1} & {\bf 58.5} \\
\midrule 
{\textbf{\model} (ours)} & {86.6} & {73.3} & {76.7} & {\bf 11.8} & {59.0} & {47.8} & {39.5} & {73.9} & {\bf 84.6} & {47.8} \\ 
\bottomrule 
\end{tabular}
\caption{Results (un-weighted by execution success) under \textit{high-level instructions} scenario.}
\label{tab:results-hl-unweighted}
\end{table*}




\section{Perceptual Analysis}
\label{app:user_eval}
In this section, we provide perceptual analysis details. We build a google doc and ask the user to score each slide from 1-5 (1 is the worst and 5 is the best), as shown in \autoref{list:user_prompt}.
\begin{figure*}[htbp]
\lstset{
    language=Python,
    basicstyle=\footnotesize\ttfamily,
    keywordstyle=\color{blue},
    stringstyle=\color{red},
    commentstyle=\color{gray},
    breaklines=true,
}
\begin{lstlisting}
"""
Please score each slide from 1-5 based on your preference to use this slide in a real presentation. 5 is the best, 1 is the worst. 

Carefully reading each slide's content before ranking. 
"""
\end{lstlisting}
\vspace{-4mm}
\caption{\textbf{Instruction we used for the perceptual evaluation.}}
\label{list:user_prompt}
\end{figure*}

\begin{figure}[htbp]
    \centering
    \includegraphics[width=0.9\linewidth]{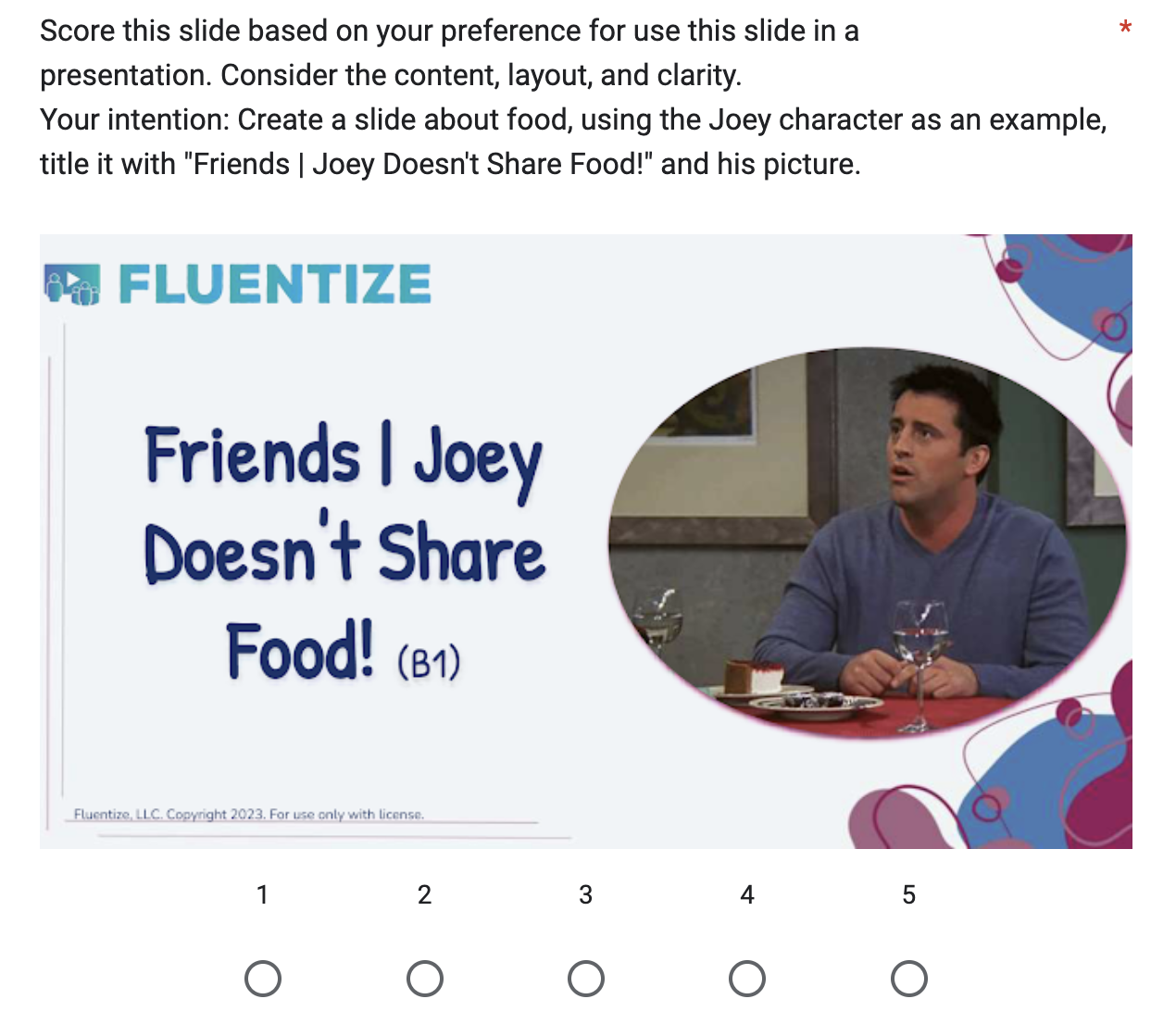}
    \caption{\textbf{An example of the perceptual analysis question.} We ask the human to score the quality of the slide from 1-5.}
    \label{fig:user_example}
\end{figure}
An example of the question is shown in \autoref{fig:user_example}.

\end{document}